\documentclass[english]{article}
\usepackage[T1]{fontenc}
\usepackage[latin9]{inputenc}
\usepackage{babel}
\usepackage{array}
\usepackage{varioref}
\usepackage{float}
\usepackage{amssymb}
\usepackage{graphicx}
\usepackage[numbers]{natbib}

\makeatletter

\newcommand{\noun}[1]{\textsc{#1}}
\providecommand{\tabularnewline}{\\}
\newcommand{\lyxdot}{.}

\floatstyle{ruled}
\newfloat{algorithm}{tbp}{loa}
\providecommand{\algorithmname}{Algorithm}
\floatname{algorithm}{\protect\algorithmname}

\usepackage{nips15submit_e,times}
\usepackage{url}
\usepackage{algorithm,algorithmicx,algpseudocode}
\usepackage{hyperref}

\author{
Pascal Vincent \\
Département d'Informatique et de Recherche Opérationnelle\\
Université de Montréal\\
Montréal, Québec, CANADA \\
and CIFAR \\
\texttt{vincentp@iro.umontreal.ca} \\
\And
Alexandre de Brébisson \\
Département d'Informatique et de Recherche Opérationnelle\\
Université de Montréal\\
Montréal, Québec, CANADA \\
\texttt{alexandre.de.brebisson@umontreal.ca} \\
\And
Xavier Bouthillier \\
Département d'Informatique et de Recherche Opérationnelle\\
Université de Montréal\\
Montréal, Québec, CANADA \\
\texttt{xavier.bouthillier@iumontreal.ca} 
}

\usepackage{enumitem}
\setlist{nolistsep}
\nipsfinalcopy 

\makeatother

\begin{document}

\title{Efficient Exact Gradient Update for training Deep Networks with Very
Large Sparse Targets\\
{[} Technical report {]}}
\maketitle
\begin{abstract}
An important class of problems involves training deep neural networks
with sparse prediction \emph{targets} of very high dimension $D$.
These occur naturally in e.g. neural language models or the learning
of word-embeddings, often posed as predicting the probability of next
words among a vocabulary of size $D$ (e.g. $200\,000$). Computing
the equally large, but typically non-sparse {\normalsize{}$D$}-dimensional
output vector from a last hidden layer of reasonable dimension $d$
(e.g. $500$) incurs a prohibitive $O(Dd)$ computational cost \emph{for
each example}, as does updating the {\normalsize{}$D\times d$} output
weight matrix and computing the gradient needed for backpropagation
to previous layers. While efficient handling of large sparse network
inputs is trivial, the case of large sparse \emph{targets} is not,
and has thus so far been sidestepped with approximate alternatives
such as hierarchical softmax or sampling-based approximations during
training. In this work we develop an original algorithmic approach
which, for a family of loss functions that includes squared error
and spherical softmax, can compute the \emph{exact} loss, gradient
update for the output weights, and gradient for backpropagation, all
in $O(d^{2})$ per example instead of $O(Dd)$, remarkably without
ever computing the $D$-dimensional output. The proposed algorithm
yields a speedup of $\frac{D}{4d}$, i.e. two orders of magnitude
for typical sizes, for that critical part of the computations that
often dominates the training time in this kind of network architecture.
\end{abstract}

\section{Introduction}

Many modern applications of neural networks have to deal with data
represented, or representable, as very large sparse vectors. Such
representations arise in natural language related tasks, where the
dimension $D$ of that vector is typically (a multiple of) the size
of the vocabulary, but also in the sparse user-item matrices of collaborative-filtering
applications. It is trivial to handle very large sparse inputs to
a neural network in a computationally efficient manner: the forward
propagation and update to the input weight matrix after backpropagation
are correspondingly sparse. By contrast, training with very large
sparse prediction \emph{targets} is problematic: even if the target
is sparse, the computation of the equally large network output and
the corresponding gradient update to the huge output weight matrix
are \emph{not sparse} and thus computationally prohibitive. This has
been a practical problem ever since \citet{BenDucVin01} first proposed
using a neural network for learning a language model, in which case
the computed output vector represents the probability of the next
word and is the size of the considered vocabulary, which is becoming
increasingly large in modern applications \citep{collobert:2011b}.
Several approaches have been proposed to attempt to address this difficulty
essentially by sidestepping it. They fall in two categories:
\begin{itemize}
\item \emph{Sampling or selection based approximations} consider and compute
only a tiny fraction of the output's dimensions sampled at random
or heuristically chosen. The reconstruction sampling of \citet{Dauphin2011},
the efficient use of biased importance sampling in \citet{Jean-et-al-ACL2015},
the use of Noise Contrastive Estimation \citep{Gutmann+Hyvarinen-2010}
in \citet{Mnih2013} and \citet{Mikolov-et-al-NIPS2013} all fall
under this category. As does the more recent use of approximate Maximum
Inner Product Search based on Locality Sensitive Hashing techniques\citep{NIPS2014_5329,arxiv1412.7479}
to select a good candidate subset. 
\item \emph{Hierarchical softmax} \citep{Morin+al-2005,Mikolov-et-al-NIPS2013}
imposes a heuristically defined hierarchical tree structure for the
computation of the normalized probability of the target class.
\end{itemize}
Compared to the initial problem of considering all $D$ output dimensions,
both kinds of approaches are crude approximations. In the present
work, we will instead investigate a way to actually perform the \emph{exact}
gradient update that corresponds to considering \emph{all} $D$ outputs,
but do so implicitly, in a computationally efficient manner, without
actually computing the $D$ outputs. This approach works for a relatively
restricted class of loss functions, the simplest of which is linear
output with squared error (a natural choice for sparse real-valued
regression targets). The most common choice for multiclass classification,
the \emph{softmax} loss is not part of that class, but we may use
an alternative \emph{spherical softmax}, which will also yield normalized
class probabilities. For simplicity, our presentation will focus on
squared error and on an online setting, and we only later briefly
mention its extension to minibatches and to a more general class of
loss functions.

\section{The problem}

\subsection{Problem definition and setup}

We are concerned with gradient-descent based training of a deep feed-forward
neural network with target vectors of very high dimension $D$ (e.g.
$D=200\,000$) but that are sparse, i.e. a comparatively small number,
at most $K\ll D$, of the elements of the target vector are non-zero.
Such a $K$-sparse vector will typically be stored and represented
compactly as $2K$ numbers corresponding to pairs \emph{(index, value)}.
A network to be trained with such targets will naturally have an equally
large output layer of dimension $D$. We can also optionally allow
the input to the network to be a similarly high dimensional sparse
vector of dimension $D_{in}$. Between the large sparse target, output,
and (optionally large sparse) input, we suppose the network's intermediate
hidden layers to be of smaller, more typically manageable, dimension
$d\ll D$ (e.g. $d=500$)\footnote{Our approach does not impose any restriction on the architecture nor
size of the hidden layers, as long as they are amenable to usual gradient
backpropagation.}.

\noindent \textbf{Mathematical notation:} Vectors are denoted using
lower-case letters, e.g. $h$, and are considered column-vectors;
corresponding row vectors are denoted with a transpose, e.g. $h^{T}$.
Matrices are denoted using upper-case letters, e.g. $W$, with $W^{T}$
the transpose of $W$. The $i^{th}$ column of $W$ is denoted $W_{i}$
, and its $i^{th}$ row $W_{:i}$ (both viewed as a column vector).
$U^{-T}=\left(U^{-1}\right)^{T}$ denotes the transpose of the inverse
of a square matrix. $\mathbf{I}_{d}$ is the $d\times d$ identity
matrix.

\begin{figure}
\centering{}\includegraphics[width=0.8\columnwidth]{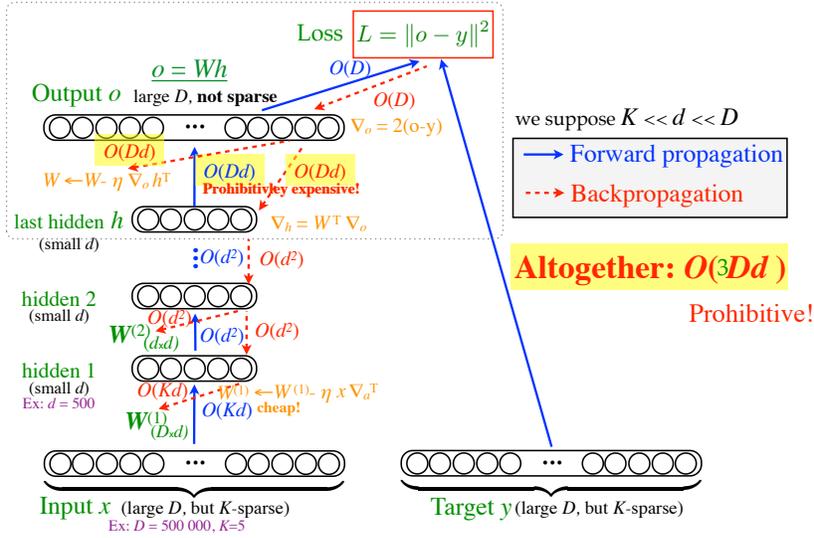}\protect\caption{\label{fig:network}The computational problem posed by very large
sparse targets. Dealing with sparse input efficiently is trivial,
with both the forward and backward propagation phases easily achieved
in $O(Kd)$. However this is not the case with large sparse targets.
They incur a prohibitive computational cost of $O(Dd)$ at the output
layer as forward propagation, gradient backpropagation and weight
update each require accessing all $D\times d$ elements of the large
output weight matrix.}
\end{figure}

\noindent \textbf{Network architecture:} We consider a standard feed
forward neural network architecture as depicted in Figure \ref{fig:network}.
An input vector $x\in\mathbb{R}^{D_{in}}$ is linearly transformed
into a linear activation $a^{(1)}=W^{(1)T}x+b^{(1)}$ through a $D_{in}\times d$
input weight matrix $W^{(1)}$ (and an optional bias vector $b^{(1)}\in\mathbb{R}^{d}$).
This is typically followed by a non-linear transformation $s$ to
yield the representation of the first hidden layer $h^{(1)}=s(a^{(1)})$.
This first hidden layer representation is then similarly transformed
through a number of subsequent non-linear layers (that can be of any
usual kind amenable to backpropagation) e.g. $h^{(k)}=s(a^{(k)})$
with $a^{(k)}=W^{(k)T}h^{(k-1)}+b^{(k)}$ until we obtain last hidden
layer representation $h=h^{(m)}$. We then obtain the final $D$-dimensional
network output as $o=Wh$ where $W$ is a $D\times d$ output weight
matrix, which will be our main focus in this work. Finally, the network's
$D$-dimensional output $o$ is compared to the $D$-dimensional target
vector $y$ associated with input $x$ using squared error, yielding
loss $L=\|o-y\|^{2}$.

\noindent \textbf{Training procedure:} This architecture is a typical
(possibly deep) multi-layer feed forward neural network architecture
with a \emph{linear output layer} and \emph{squared error loss}.
Its parameters (weight matrices and bias vectors) will be trained
by gradient descent, using gradient backpropagation \citet{Rumelhart86b-small,LeCun85,LeCun86}
to efficiently compute the gradients. The procedure is shown in Figure
\ref{fig:network}. Given an example from the training set as an \emph{(input,target)}
pair $(x,y)$, a pass of forward propagation proceeds as outlined
above, computing the hidden representation of each hidden layer in
turn based on the previous one, and finally the network's predicted
output $o$ and associated loss $L$. A pass of gradient backpropagation
then works in the opposite direction, starting from $\nabla_{o}=\frac{\partial L}{\partial o}=2(o-y)$
and propagating back the gradients $\nabla_{h^{(k)}}=\frac{\partial L}{\partial h^{(k)}}$
and $\nabla_{a^{(k)}}=\frac{\partial L}{\partial a^{(k)}}$ upstream
through the network. The corresponding gradient contributions on parameters
(weights and biases), collected along the way, are straightforward
once we have the associated $\nabla_{a^{(k)}}$. Specifically they
are $\nabla_{b^{(k)}}=\nabla_{a^{(k)}}$ and $\nabla_{W^{(k)}}=h^{(k-1)}(\nabla_{a^{(k)}})^{T}$.
Similarly for the input layer $\nabla_{W^{(1)}}=x(\nabla_{a^{(1)}})^{T}$,
and for the output layer $\nabla_{W}=(o-y)h^{T}$ . Parameters are
then updated through a gradient descent step $W^{(k)}\leftarrow W^{(k)}-\eta\nabla_{W^{(k)}}$
and $b^{(k)}\leftarrow b^{(k)}-\eta\nabla_{b^{(k)}}$, where $\eta$
is a positive learning-rate. Similarly for the output layer which
will be our main focus here: $W\leftarrow W-\eta\nabla_{W}$.

\subsection{\label{sub:easy-part}The easy part: input layer forward propagation
and weight update}

It is easy and straightforward to efficiently compute the forward
propagation, and the backpropagation and weight update part for the
\emph{input layer} when we have a very large $D_{in}$-dimensional
but $K-$sparse input vector $x$ with appropriate sparse representation.
Specifically we suppose that $x$ is represented as a pair of vectors
$u,v$ of length (at most) $K$, where $u$ contains integer indexes
and $v$ the associated real values of the elements of $x$ such that
$x_{i}=0$ if $i\notin u$, and $x_{u_{k}}=v_{k}$.
\begin{itemize}
\item \noindent \textbf{Forward propagation through the input layer:} The
sparse representation of $x$ as the positions of $K$ elements together
with their value makes it cheap to compute $W^{(1)T}x$. Even though
$W^{(1)}$ may be a huge full $D_{in}\times d$ matrix, only $K$
of its rows (those corresponding to the non-zero entries of $x$)
need to be visited and summed to compute $W^{(1)T}x$. Precisely,
with our $(u,v)$ sparse representation of $x$ this operation can
be written as$W^{(1)T}x=\sum_{k=1}^{K}v_{k}W_{:u_{k}}^{(1)}$where
each $W_{:u_{k}}^{(1)}$ is a $d$-dimensional vector, making this
an $O(Kd)$ operation rather than $O(Dd)$.
\item \textbf{Gradient and update through input layer:} Let us for now suppose
that we were able to get gradients (through backpropagation) up to
the first hidden layer activations $a^{(1)}\in\mathbb{R}^{d}$ in
the form of gradient vector $\nabla_{a^{(1)}}=\frac{\partial L}{\partial a^{(1)}}$.
The corresponding gradient-based update to input layer weights $W^{(1)}$
is simply $W^{(1)}\leftarrow W^{(1)}-\eta x(\nabla_{a^{(1)}})^{T}$.
This is a rank-one update to $W^{(1)}$. Here again, we see that only
the $K$ rows of $W^{(1)}$ associated to the (at most) $K$ non-zero
entries of $x$ need to be modified. Precisely this operation can
be written as:$W_{:u_{k}}^{(1)}\leftarrow W_{:u_{k}}^{(1)}-\eta v_{k}\nabla_{a^{(1)}}\,\,\,\forall k\in\{1,\ldots,K\}$
making this again a $O(Kd)$ operation rather than $O(Dd)$.
\end{itemize}

\subsection{\label{sub:hard-part}The hard part: output layer propagation and
weight update}

Given some network input $x$ we suppose we can compute without difficulty
through forward propagation the associated last hidden layer representation
$h\in\mathbb{R}^{d}$. From then on:
\begin{itemize}
\item Computing the final output $o=Wh$ incurs a prohibitive computational
cost of $O(Dd)$ since $W$ is a full $D\times d$ matrix. Note that
there is a-priori no reason for representation $h$ to be sparse (e.g.
with a sigmoid non-linearity) but even if it was, this would not fundamentally
change the problem since it is $D$ that is extremely large, and we
supposed $d$ reasonably sized already. Computing the residual $(\mbox{o}-t)$
and associated squared error loss $\|\mbox{o}-t\|^{2}$ incurs an
additional $O(D)$ cost.
\item The gradient on $h$ that we need to backpropagate to lower layers
is $\nabla_{h}=\frac{\partial L}{\partial h}=2W^{T}(o-y)$ which is
another $O(Dd)$ matrix-vector product.
\item Finally, when performing the corresponding output weight update $W\leftarrow W-\eta(o-y)h^{T}$
we see that it is a rank-one update that updates all $D\times d$
elements of $W$, which again incurs a prohibitive $O(Dd)$ computational
cost.
\end{itemize}
For very large $D$, all these three $O(Dd)$ operations are prohibitive,
and the fact that $y$ is sparse, seen from this perspective, doesn't
help, since neither $o$ nor $o-y$ will be sparse.

\section{A computationally efficient algorithm for performing the exact online
gradient update}

Previously proposed workarounds are approximate or use stochastic
sampling. We propose a different approach that results in the \emph{exact
same}, yet efficient gradient update, remarkably without ever having
to compute large output $o$.

\subsection{\label{sub:efficient-squared-loss}Computing the squared error loss
$L$ and the gradient with respect to $h$ efficiently}

Suppose that, we have, for a network input example $x$, computed
the last hidden representation $h\in\mathbb{R}^{d}$ through forward
propagation. The network's $D$ dimensional output $o=Wh$ is then
in principle compared to the high dimensional target $y\in\mathbb{R}^{D}$.
The corresponding squared error loss is $L=\left\Vert Wh-y\right\Vert ^{2}$.
As we saw in Section \ref{sub:hard-part}, computing it in the direct
naive way would have a prohibitive computational complexity of $O(Dd+D)=O(Dd)$
because computing output $Wh$ with a full $D\times d$ matrix $W$
and a typically non-sparse $h$ is $O(Dd)$. Similarly, to backpropagate
the gradient through the network, we need to compute the gradient
of loss $L$ with respect to last hidden layer representation $h$.
This is $\nabla_{h}=\frac{\partial L}{\partial h}=\frac{\partial\left\Vert Wh-y\right\Vert ^{2}}{\partial h}=2W^{T}(Wh-y)$.
So again, if we were to compute it directly in this manner, the computational
complexity would be a prohibitive $O(Dd)$. \textbf{Provided we have
maintained an up-to-date matrix $Q=W^{T}W$}, which is of reasonable
size $d\times d$ and can be cheaply maintained as we will see in
Section \ref{sub:bookkeeping}, we can rewrite these two operations
so as to perform them in $O(d^{2})$:

\begin{tabular}{c|c}
\multicolumn{1}{c}{\textbf{Loss computation:}} & \textbf{Gradient on $h$:}\tabularnewline
\begin{minipage}[t]{0.5\columnwidth}%
\begin{eqnarray}
L & = & \|\overbrace{Wh}^{O(Dd)}-y\|^{2}\nonumber \\
 & = & \left(Wh-y\right)^{T}\left(Wh-y\right)\nonumber \\
 & = & h^{T}W^{T}Wh-y^{T}Wh-h^{T}W^{T}y+y^{T}y\nonumber \\
 & = & h^{T}Qh-2h^{T}(W^{T}y)+y^{T}y\nonumber \\
 & = & h^{T}(\underbrace{Qh}_{O(d^{2})}-2\underbrace{W^{T}y}_{O(Kd)})+\underbrace{y^{T}y}_{O(K)}\label{eq:L}
\end{eqnarray}
\end{minipage} & %
\begin{minipage}[t]{0.5\columnwidth}%
\begin{eqnarray}
\nabla_{h}=\frac{\partial L}{\partial h} & = & \frac{\partial\|Wh-y\|^{2}}{\partial h}\nonumber \\
 & = & 2W^{T}(Wh-y)\nonumber \\
 & = & 2\left(W^{T}Wh-W^{T}y\right)\nonumber \\
 & = & 2(\underbrace{Qh}_{O(d^{2})}-\underbrace{W^{T}y}_{O(Kd)})\label{eq:gradH}
\end{eqnarray}
\end{minipage}\tabularnewline
\end{tabular}

The terms in $O(Kd)$ and $O(K)$ are due to leveraging the $K$-sparse
representation of target vector $y$. With $K\ll D$ and $d\ll D$,
we get altogether a computational cost of $O(d^{2})$ which can be
several orders of magnitude cheaper than the prohibitive $O(Dd)$
of the direct approach.

\subsection{Efficient gradient update of $W$}

The gradient of the squared error loss with respect to output layer
weight matrix $W$ is $\frac{\partial L}{\partial W}=\frac{\partial\left\Vert Wh-y\right\Vert ^{2}}{\partial W}=2(Wh-y)h^{T}$.
And the corresponding gradient descent update to $W$ would be $W_{new}\leftarrow W-2\eta(Wh-y)h^{T}$,
where $\eta$ is a positive learning rate. Again, computed in this
manner, this induces a prohibitive $O(Dd)$ computational complexity,
both to compute output and residual $Wh-y$, and then to update all
the $Dd$ elements of $W$ (since generally neither $Wh-y$ nor $h$
will be sparse). All $D\times d$ elements of $W$ must be accessed
during this update. On the surface this seems hopeless. But we will
now see how we can achieve the \emph{exact} same update on $W$ in
$O(d^{2})$. The trick is to represent $W$ \emph{implicitly} as the
factorization $\underbrace{W}_{D\times d}=\underbrace{V}_{D\times d}\underbrace{U}_{d\times d}$and
update $U$ and $V$ instead:

\begin{eqnarray}
\mathbf{a)}\,\,U_{new} & = & U-2\eta(Uh)h^{T}\label{eq:updateU}\\
\mathbf{b)}\,\,V_{new} & = & V+2\eta y(U_{new}^{-T}h){}^{T}\grave{u}\label{eq:updateV}
\end{eqnarray}

This results in \emph{implicitly} updating $W$ as we did \emph{explicitly}
in the naive approach as we now prove:

\begin{eqnarray*}
V_{new}U_{new} & = & (V+2\eta y(U_{new}^{-T}h){}^{T})\,U_{new}\\
 & = & VU_{new}+2\eta y(U_{new}^{-T}h){}^{T}U_{new}\\
 & = & VU_{new}+2\eta yh^{T}U_{new}^{-1}U_{new}\\
 & = & V(U-2\eta(Uh)h^{T})+2\eta yh^{T}(U_{new}^{-1}U_{new})\\
 & = & VU-2\eta VUhh^{T}+2\eta yh^{T}\\
 & = & VU-2\eta(VUh-y)h^{T}\\
 & = & W-2\eta(Wh-y)^{T}h^{T}\\
 & = & W_{new}
\end{eqnarray*}

We see that the update of $U$ in Eq.~\ref{eq:updateU} is a simple
$O(d^{2})$ operation. Following this simple rank-one update to $U$,
we can use the Sherman-Morrison formula to derive the corresponding
rank-one update to $U^{-T}$ which will also be $O(d^{2})$:

\begin{eqnarray}
U_{new}^{-T} & = & U^{-T}+\frac{2\eta}{1-2\eta\left\Vert h\right\Vert ^{2}}(U^{-T}h)h^{T}\label{eq:updateU-T}
\end{eqnarray}
It is then easy to compute the $U_{new}^{-T}h$, an $O(d^{2})$ operation
needed in Eq.~\ref{eq:updateV}. The ensuing rank-one update of $V$
in Eq~\ref{eq:updateV}, thanks to the $K$-sparsity of $y$ is only
$O(Kd)$: only the$K$ rows $V$ associated to non-zero elements in
$y$ are accessed and updated, sited of \emph{all} $D$ rows of $W$
we had to modify in the naive update! 

Note that with the factored representation of $W$ as $VU$, we only
have $W$ implicitly, so the $W^{T}y$ terms that entered in the computation
of $L$ and $\nabla_{h}$ in the previous paragraph need to be adapted
slightly as $\hat{y}=W^{T}y=U^{T}(V^{T}y)$, which becomes $O(d^{2}+Kd)$
rather than $O(Kd)$ in computational complexity. But this doesn't
change the overall $O(d^{2})$ complexity of these computations.

\subsection{\label{sub:bookkeeping}Bookkeeping: keeping an up-to-date $Q$ and
$U^{-T}$}

We have already seen, in Eq.~\ref{eq:updateU-T}, how we can cheaply
maintain an up-to-date $U^{-T}$ following our update of $U$. Similarly,
following our updates to $U$ and $V$, we need to keep an up-to-date
$Q=W^{T}W$ which is needed to efficiently compute the loss $L$ (Eq.~\ref{eq:L})
and gradient $\nabla_{h}$ (Eq.~\ref{eq:gradH}). We have shown that
updates to $U$ and $V$ in equations~\ref{eq:updateU} and \ref{eq:updateV}
are equivalent to implicitly updating $W$ as $W_{new}\leftarrow W-2\eta(Wh-y)h^{T}$,
and this translates into the following update to $Q=W^{T}W$:

\begin{eqnarray}
\hat{z} & = & Qh-U^{T}(V^{T}y)\nonumber \\
Q_{new} & = & Q-2\eta\left(h\hat{z}^{T}+\hat{z}h^{T}\right)+(4\eta^{2}L)hh^{T}\label{eq:updateQ}
\end{eqnarray}

The proof is straightforward but due to space constraints we put it
in supplementary material. One can see that this last bookkeeping
operation also has a $O(d^{2})$ computational complexity.

\subsubsection*{$ $}

\subsection{Putting it all together: detailed algorithm and expected benefits}

We have seen that we can efficiently compute cost $L$, gradient with
respect to $h$ (to be later backpropagated further) as well as updating
$U$ and $V$ and performing the bookkeeping for $U^{-T}$ and $Q$.
Algorithm~ \ref{alg:online} describes the detailed algorithmic steps
that we put together from the equations derived above. Having $K\ll d\ll D$
we see that the proposed algorithm requires $O(d^{2})$ operations,
whereas the standard approach required $O(Dd)$ operations. If we
take $K\approx d$ , we may state more precisely that the proposed
algorithm, for computing the loss and the gradient updates will require
roughly $12d^{2}$ operations whereas the standard approach required
roughly $3Dd$ operations. So overall the proposed algorithm change
corresponds to a computational speedup by a factor of $\frac{D}{4d}$.
For $D=200\,000$ and $d=500$ the expected speedup is thus \textbf{100}.
Note that the advantage is not only in \emph{computational} complexity,
but also in \emph{memory access}. For each example, the standard approach
needs to access and change all $D\times d$ elements of matrix $W$,
whereas the proposed approach only accesses the much smaller number
$K\times d$ elements of $V$ as well as the three $d\times d$ matrices
$U$, $U^{-T}$, and $Q$. So overall we have a \textbf{substantially
faster algorithm}, which, while doing so \emph{implicitly}, will nevertheless
perform the \emph{exact same} gradient update as the standard approach.
We want to emphasize here that our approach is completely different
from simply chaining 2 linear layers $U$ and $V$ and performing
ordinary gradient descent updates on these: this would result in the
same prohibitive computational complexity as the standard approach,
and such ordinary separate gradient updates to $U$and $V$ would
not be equivalent to the ordinary gradient update to $W=VU$.

\begin{algorithm}
\begin{centering}
\protect\caption{\label{alg:online}Efficient computation of cost $L$, gradient $h$,
and update to parameters $U$ and $V$}
\begin{tabular}{|>{\raggedright}p{0.05\textwidth}|>{\raggedright}p{0.3\textwidth}|>{\centering}p{0.15\textwidth}|>{\centering}p{0.15\textwidth}|}
\hline 
\textbf{Step \#} & \textbf{Operation} & \textbf{Computational complexity} & \textbf{Number of multiply-adds}\tabularnewline
\hline 
\hline 
1: & $\hat{h}=Qh$  & $O(d^{2})$ & $d^{2}$\tabularnewline
\hline 
2: & $\hat{y}=U^{T}(V^{T}y)$ & $O(Kd+d^{2})$ & $Kd+d^{2}$\tabularnewline
\hline 
3: & $\hat{z}=\hat{h}-\hat{y}$ & $O(d)$ & $d$\tabularnewline
\hline 
4: & $\nabla_{h}=2\hat{z}$ & $O(d)$ & $d$\tabularnewline
\hline 
5: & $L=h^{T}\hat{h}-2h^{T}\hat{y}+y^{T}y$ & $O(2d+K)$ & $2d+K+1$\tabularnewline
\hline 
6: & $U_{new}=U-2\eta(Uh)h^{T}$ & $O(d^{2})$ & $2d^{2}+d$\tabularnewline
\hline 
7: & $U_{new}^{-T}=U^{-T}+\frac{2\eta}{1-2\eta\left\Vert h\right\Vert ^{2}}(U^{-T}h)h^{T}$ & $O(d^{2})$ & $2d^{2}+2d+3$\tabularnewline
\hline 
8: & $V_{new}=V+2\eta y(U_{new}^{-T}h){}^{T}$ & $O(d^{2}+Kd)$ & $d^{2}+K+Kd$\tabularnewline
\hline 
9: & $Q_{new}=Q-2\eta\left(h\hat{z}^{T}+\hat{z}h^{T}\right)+(4\eta^{2}L)hh^{T}$ & $O(d^{2})$ & $4+2d+3d^{2}$\tabularnewline
\hline 
 & \textbf{Altogether:} & $O(d^{2})$

provided $K<d\ll D$ & $\approx12d^{2}$ elementary operations\tabularnewline
\hline 
\end{tabular}
\par\end{centering}

\end{algorithm}

\subsection{Controlling numerical stability and extension to the minibatch case}

The update of $U$ in Equation \ref{eq:updateU} may over time lead
$U$ to become ill-conditioned. To prevent this, we regularly (every
100 updates) monitor its conditioning number\footnote{Largest and smallest singular value can be computed with an SVD or
using the power iteration method.}. If either the smallest or largest singular value moves outside an
acceptable range, we bring it back to 1 by doing an appropriate rank-1
update to $V$ (which costs $Dd$ operations, but is only done rarely).
Our algorithm can also be straightforwardly extended to the minibatch
case (the derivations are given in the supplementary material section)
and yields the same theoretical speedup factor with respect to the
standard naive approach. But one needs to be careful in order to keep
the computation of $U^{-T}h$ reasonably efficient: depending on the
size of the minibatch $m$, it may be more efficient to solve the
corresponding linear equation for each minibatch from scratch rather
than updating $U^{-T}$with the Woodbury equation (which generalizes
the Sheman-Morrison formula for $m>1$).

\subsection{Generalization to a broader class of loss functions}

The approach that we detailed for linear output and squared error
can easily be extended to slightly more exotic loss functions: basically
any loss function that can be expressed using only the $o_{c}$ associated
to non-zero $y_{c}$ and $\|o\|^{2}=\sum_{j}o_{j}^{2}$ the squared
norm of the whole output vector, which we can compute cheaply. This
family of loss functions does not include the standard log of softmax,
but includes the so-called \emph{spherical softmax}: $\log\frac{(o_{c}+\epsilon)^{2}}{\sum_{j}(o_{j}+\epsilon)^{2}}$
(where $c$ is the correct class label). It remains to be seen in
practice how this approach performs computationally, and whether we
lose something due to using this more limited family of loss functions.

\section{Experimental validation}

\begin{figure}
\includegraphics[width=0.5\textwidth]{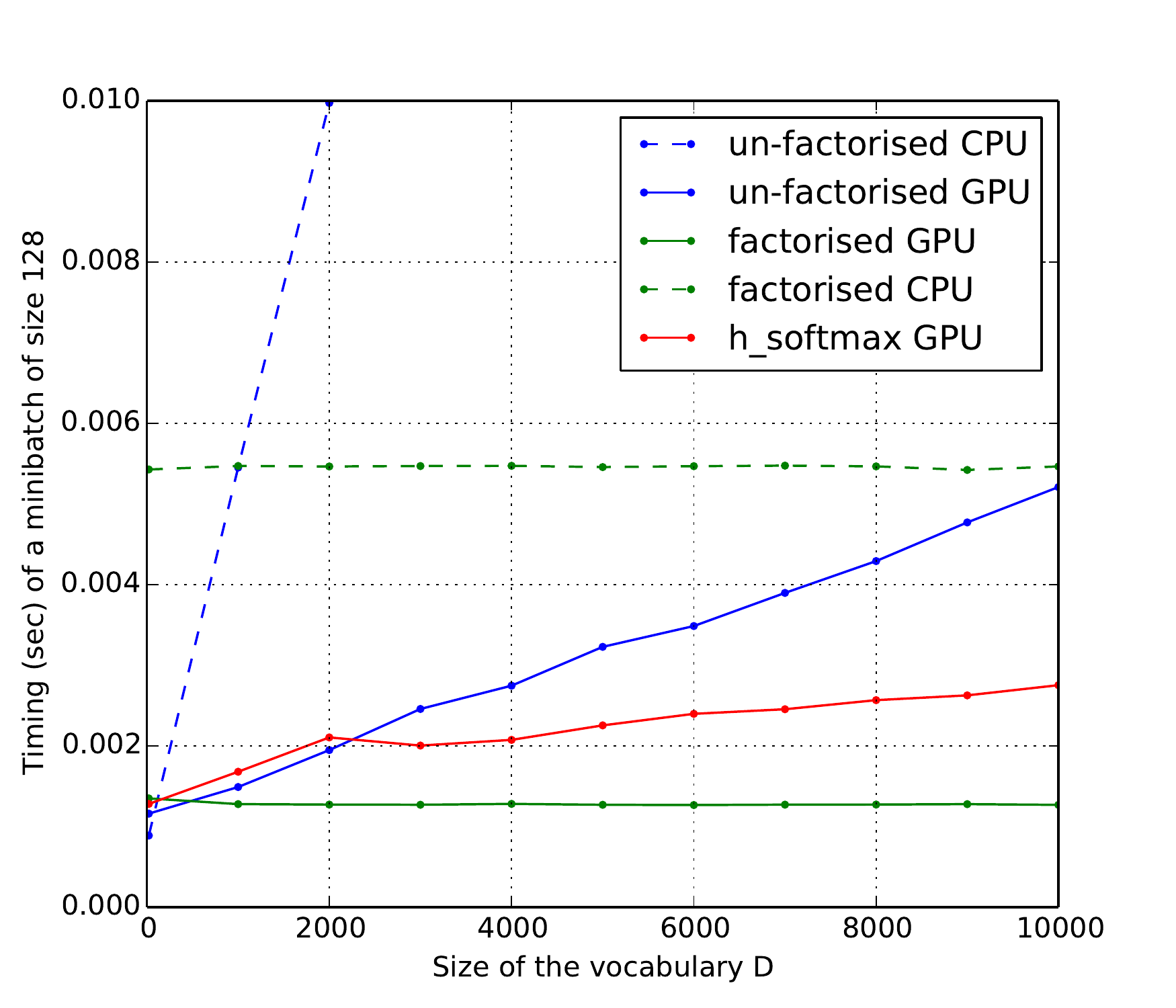}\includegraphics[width=0.5\textwidth]{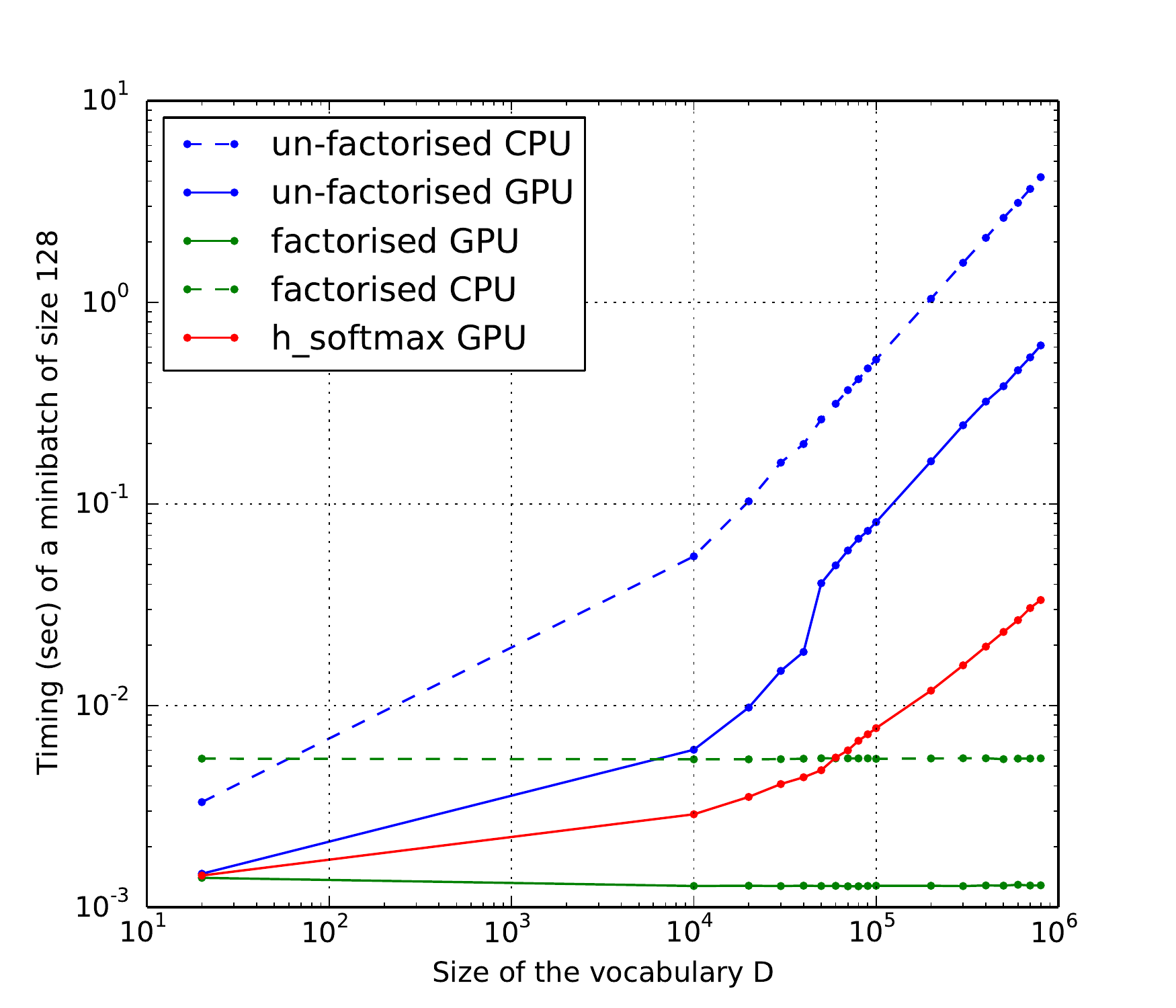}

\protect\caption{\label{fig:Timing-of-different}Timing of different algorithms. Time
taken by forward and backward propagations in the output layer, including
weight update, on a minibatch of size 128 for different sizes of vocabulary
D on both CPU and GPU. The input size d is fixed to 300. The Timing
of a 2 layer hierarchical softmax efficient GPU implementation (h\_softmax)
is also provided for comparison. Right plot is in log-log scale. As
expected, the timings of factorized versions are independent of the
size of the vocabulary.}

\end{figure}

\begin{figure}
\includegraphics[width=0.5\textwidth]{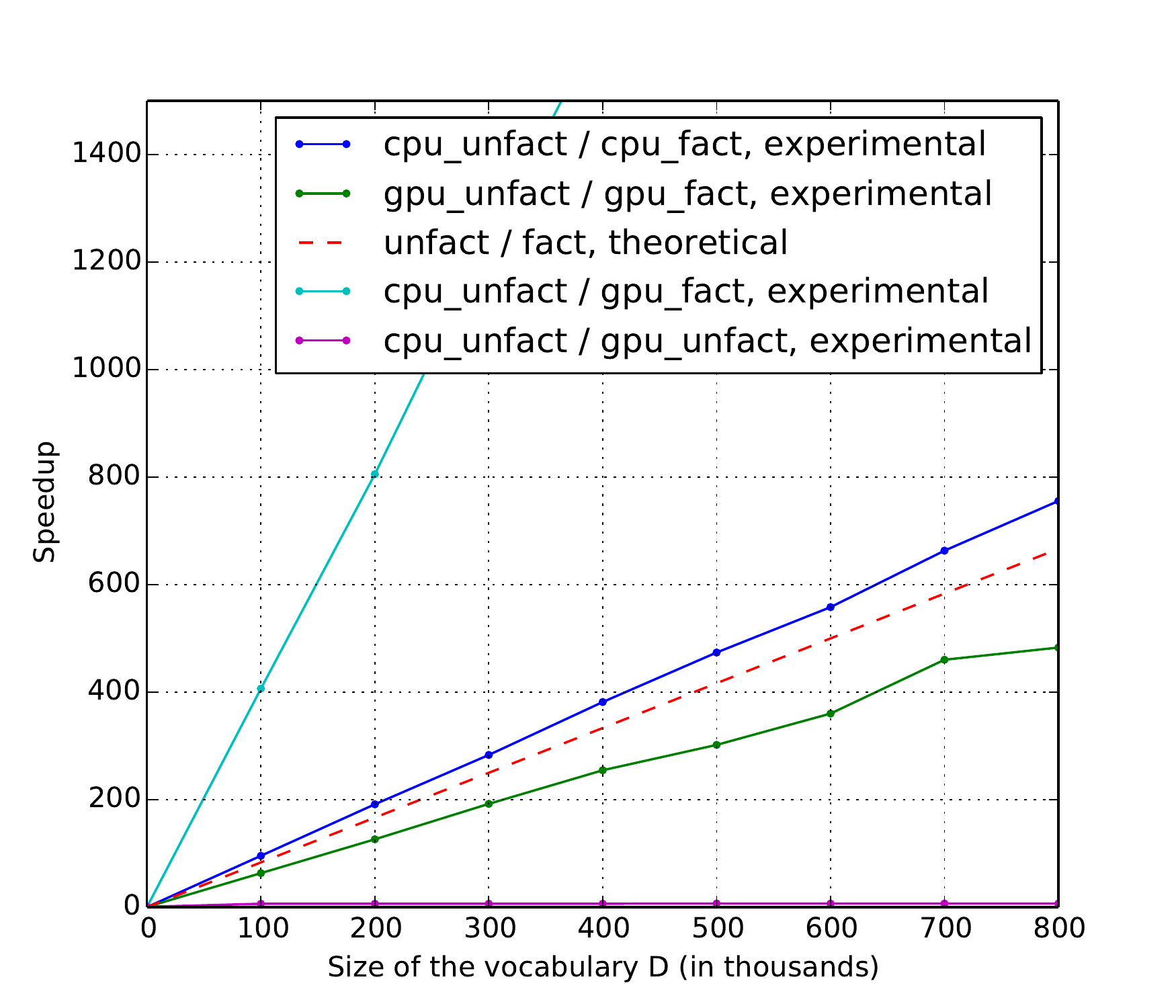}\includegraphics[width=0.5\textwidth]{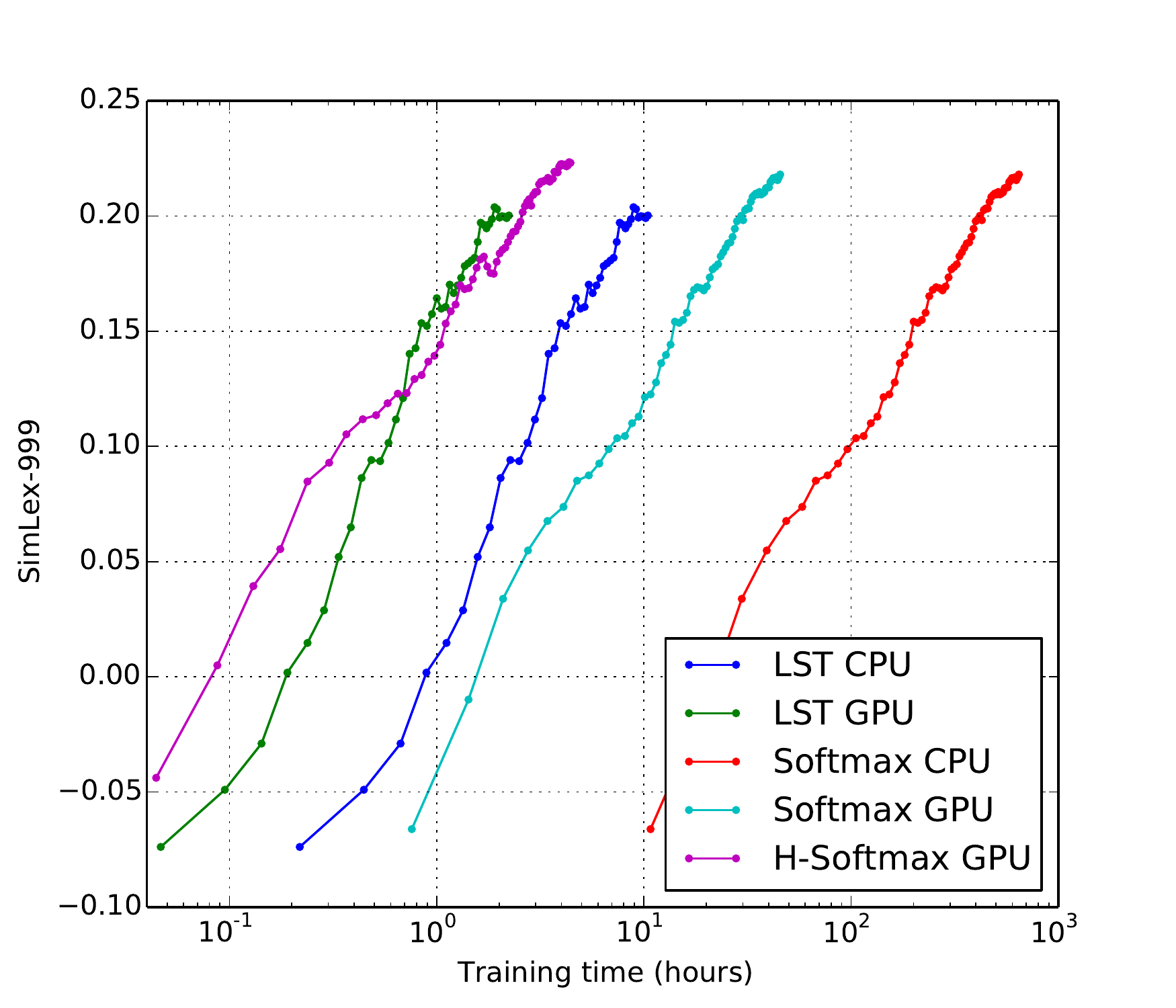}\protect\caption{\textbf{\label{fig:speedup}Left:} Practical and theoretical speedups
for different sizes of vocabulary D and fixed input size d=300. The
practical unfact / fact speedup is similar to the theoretical one.
\textbf{Right}: Evolution of the Simlex-999 score obtained with different
models as a function of training time (CPU softmax times were extrapolated
from fewer iterations). Softmax models are zero hidden-layer models,
while our large sparse target (LST) models have two hidden layers.
These were the best architectures retained in both cases (surprisingly
the softmax models with hidden layers performed no better on this
task). The extra non-linear layers in LST may help compensate for
the lack of a softmax. LST models converge to slightly lower scores
at similar speed as the hierarchical softmax model but significantly
faster than softmax models. }
\end{figure}

We implemented both a CPU version using \emph{blas} and a parallel
GPU (Cuda) version using \emph{cublas} of the proposed algorithm\footnote{Open source code will be released upon official publication of this
research.}. We evaluated the GPU and CPU implementations by training word embeddings
with simple neural language models, in which a probability map of
the next word given its preceding n-gram is learned by a neural network.
We used a Nvidia Titan Black GPU and a i7-4820K @ 3.70GHz CPU and
ran experiments on the one billion word dataset\citep{DBLP:conf/interspeech/ChelbaMSGBKR14},
which is composed of 0.8 billions words belonging to a vocabulary
of 0.8 millions words. We evaluated the resulting word embeddings
with the recently introduced Simlex-999 score \citep{DBLP:journals/corr/HillRK14},
which measures the similarity between words. We also compared our
approach to unfactorised versions and to a two-layer hierarchical
softmax. Figure \ref{fig:Timing-of-different} and \ref{fig:speedup}
(left) illustrate the practical speedup of our approach for the output
layer only. Figure \ref{fig:speedup}(right) shows that the LST (Large
Sparse Target) models are much faster to train than the softmax models
and converge to only slightly lower Simlex-999 scores. Table \ref{tab:Speedups-with-respect}
summarizes the speedups for the different output layers we tried,
both on CPU and GPU. We also emprically verified that our proposed
factored algorithm learns the exact same model weights $(VU)$ as
the corresponding naive unfactored algorithm's $W$, as it theoretically
should (up to negligible numerical precision differences), and followed
the exact same learning curves (as a function of number of iterations,
not time!). 

\begin{table}
\centering{}\protect\caption{\label{tab:Speedups-with-respect}Speedups with respect to the baseline
naive model on CPU, for a minibatch of 128 and the whole vocabulary
of D = 793471 words. This is a two hidden layer model with 300 neurons
on all its layers (so d = 300). }
\begin{tabular}{|c|c|c|}
\hline 
\textbf{Model} & \textbf{output layer only speedup} & \textbf{whole model speedup}\tabularnewline
\hline 
\hline 
cpu unfactorised (naive) & 1 & 1\tabularnewline
\hline 
gpu unfactorised (naive) & 6.8 & 4.7\tabularnewline
\hline 
gpu hierarchical softmax & 125.2 & 178.1\tabularnewline
\hline 
cpu factorised & 763.3 & 501\tabularnewline
\hline 
gpu factorised & 3257.3 & 1852.3\tabularnewline
\hline 
\end{tabular}
\end{table}

\section{Conclusion and future work}

We introduced a new algorithmic approach to efficiently compute the
\emph{exact} gradient updates for training deep networks with very
large sparse targets. Remarkably the complexity of the algorithm is
independent of the target size, which allows tackling very large problems.
Our CPU and GPU implementation yield similar speedups to the theoretical
one and can thus be used in practical applications, which could be
explored in further work. In particular, neural language models seem
good candidates. But it remains unclear how using a loss function
other than \emph{log-softmax} may affect the quality of the resulting
word embeddingsm and further research should be carried out in this
direction. Extensions of the approach to other possible losses than
the simple squared error should also be empirically investigated in
this light, in particular \emph{log-spherical-softmax}.

\section*{Acknowledgements}

We would like to thank the developers of Theano \citep{bergstra+al:2010-scipy,Bastien-Theano-2012}
and Blocks \citep{blocksfuel}. 

This research is supported by NSERC and Ubisoft.

\small

\bibliographystyle{unsrtnat_without_url}
\bibliography{strings,strings-shorter,ml,aigaion-shorter,large_sparse}

\normalsize

\clearpage{}

\appendix

\part*{Appendix}

\section{Minibatch version of the algorithm}

The algorithm we derived for online gradient is relatively straightforward
to extend to the case of minibatches containing $m$ examples, and
will still yield the same theoretical speedup factor with respect
to the standard naive approach. One may want to be careful in order
to keep the computation of $U^{-T}h$ (or ore precisely $U^{-T}H$
in the minibatch case) reasonably efficient. In the minibatch version
presented below, we update $U^{-T}$ based on the Woodbury equation
(which generalizes the Sheman-Morrison formula for $m>1$ and involves
inverting an $m\times m$ matrix). But depending on the size of the
minibatch $m$, it may become more efficient to solve the corresponding
linear equations for each minibatch from scratch every time, rather
than inverting that $m\times m$ matrix. In which case we won't need
to maintain an $U^{-T}$ at all.

\begin{algorithm}[H]

\protect\caption{Minibatch version of the update algorithm}

\subsection*{Initialization}
\begin{itemize}
\item we can initialize $D\times d$ matrix $V$ randomly as we would have
initialized $W$ so that we initially have $V=W$.\\
Alternatively we can initialize $V$ to 0 (there won't be symmetry
breaking issues with having $W$ initially be 0 provided the other
layers are initialized randomly, since varying inputs and targets
will naturally break symmetry for the output layer)
\item initialize $Q\leftarrow V^{T}V$ (or more cheaply initialize $Q\leftarrow0$
if we have initialized $V$ to 0).
\item we initialize $U$ to the identity: $U\leftarrow\mathbf{I}_{d}$ so
that, trivially, we initially have $VU=W$.
\item initialize $U^{-T}\leftarrow\mathbf{I}_{d}$
\end{itemize}

\subsection*{Update}

We suppose we receive $m$ target vectors in the $m$ columns of sparse
matrix $Y$, and corresponding $m$ hidden representations in the
$m$ columns of matrix $H$. 

\begin{tabular}{|>{\raggedright}p{0.05\textwidth}|>{\raggedright}p{0.4\textwidth}|>{\centering}p{0.15\textwidth}|>{\raggedright}p{0.3\textwidth}|}
\hline 
Step \# & Operation & Computation complexity & Computational complexity with the multiplicative factor left in.\tabularnewline
\hline 
\hline 
1: & $\hat{H}=QH$  & $O(md^{2})$ & $O(md^{2})$\tabularnewline
\hline 
2: & $\hat{Y}=U^{T}(V^{T}Y)$ & $O(mKd+md^{2})$ & $O(mKd+md^{2})$\tabularnewline
\hline 
3: & $\hat{Z}=\hat{H}-\hat{Y}$ & $O(md)$ & $O(md)$\tabularnewline
\hline 
4: & $\nabla_{H}=2\hat{Z}$ & $O(md)$ & $O(md)$\tabularnewline
\hline 
5: & $M=H^{T}\hat{Z}-\hat{Y}^{T}H+Y^{T}Y$ or alternatively $M=H^{T}\hat{H}-(\hat{Y}^{T}H+H^{T}\hat{Y})+Y^{T}Y$ & $O(m^{2}d+m^{2}K)$ & $O(2m^{2}d+m^{2}K)$\tabularnewline
\hline 
6: & $L=\mathrm{Tr}(M)$ & $O(m)$ & $O(m)$\tabularnewline
\hline 
7: & $U_{new}=U-2\eta(UH)H^{T}$ & $O(md^{2})$ & $O(2md^{2})$\tabularnewline
\hline 
8: & $U_{new}^{-T}=U^{-T}-(U^{-T}H)\left((H^{T}H-\frac{1}{2\eta}\mathbf{I}_{m})^{-1}H^{T}\right)$ & $O(m^{2}d+m^{3}+md^{2})$ & $O(2m^{2}d+m^{3}+2md^{2})$\tabularnewline
\hline 
9: & $V_{new}=V+2\eta Y(U_{new}^{-T}H){}^{T}$ & $O(md^{2}+mKd)$ & $O(md^{2}+mKd)$\tabularnewline
\hline 
10: & $Q_{new}=Q-2\eta\left(H\hat{Z}^{T}+\hat{Z}H^{T}\right)+4\eta^{2}(HM)H^{T}$ & $O(md^{2}+dm^{2})$ & $O(3md^{2}+m^{2}d)$\tabularnewline
\hline 
\end{tabular}

\end{algorithm}

\newpage{}

\section{Detailed proof for computation of update of $Q$}

Update to $Q$ corresponds to $W_{new}\leftarrow W-2\eta(WH-Y)H^{T}$

We will use the following precomputed quantities: $Q=W^{T}W$, $\hat{H}=QH$
and $\hat{Y}=W^{T}Y=U^{T}(V^{T}Y)$ and $\hat{Z}=\hat{H}-\hat{Y}$.

\begin{eqnarray*}
Q_{new} & = & W_{new}^{T}W_{new}\\
 & = & \left(W-2\eta(WH-Y)H^{T}\right)^{T}\left(W-2\eta(WH-Y)H^{T}\right)\\
 & = & W^{T}W-2\eta H(WH-Y)^{T}W-2\eta W^{T}(WH-Y)H^{T}\\
 &  & +4\eta^{2}H(WH-Y)^{T}(WH-Y)H^{T}\\
 & = & Q-2\eta\left(HH^{T}W^{T}W-HY^{T}W\right)-2\eta\left(W^{T}WHH^{T}-W^{T}YH^{T}\right)\\
 &  & +4\eta^{2}H(H^{T}W^{T}WH-H^{T}W^{T}Y-Y^{T}WH+Y^{T}Y)H^{T}\\
 & = & Q-2\eta\left(HH^{T}Q-H(W^{T}Y)^{T}\right)-2\eta\left(QHH^{T}-(W^{T}Y)H^{T}\right)\\
 &  & +4\eta^{2}H(H^{T}QH-H^{T}(W^{T}Y)-(W^{T}Y)^{T}H+Y^{T}Y)H^{T}\\
 & = & Q-2\eta\left(H\hat{H}^{T}-H\hat{Y}^{T}+\hat{H}H^{T}-\hat{Y}H^{T}\right)\\
 &  & +4\eta^{2}H(H^{T}\hat{H}-H^{T}\hat{Y}-\hat{Y}^{T}H+Y^{T}Y)H^{T}\\
 & = & Q-2\eta\left(H(\hat{H}-\hat{Y})^{T}+(\hat{H}-\hat{Y})H^{T}\right)+4\eta^{2}H(H^{T}(\hat{H}-\hat{Y})-\hat{Y}^{T}H+Y^{T}Y)H^{T}\\
 & = & Q-2\eta\left(H\hat{Z}^{T}+\hat{Z}H^{T}\right)+4\eta^{2}H\underbrace{\left(H^{T}\hat{Z}-\hat{Y}^{T}H+Y^{T}Y\right)}_{M}H^{T}
\end{eqnarray*}

This is what is listed as step 10 of the above minibatch algorithm.

In the online case, this becomes:

\begin{eqnarray*}
Q_{new} & = & Q-2\eta\left(h\hat{z}^{T}+\hat{z}h^{T}\right)+4\eta^{2}\left(h^{T}\hat{z}-\hat{y}^{T}h+y^{T}y\right)hh^{T}\\
 & = & Q-2\eta\left(h\hat{z}^{T}+\hat{z}h^{T}\right)+4\eta^{2}\left(h^{T}\hat{h}-h^{T}\hat{y}-\hat{y}^{T}h+y^{T}y\right)hh^{T}\\
 & = & Q-2\eta\left(h\hat{z}^{T}+\hat{z}h^{T}\right)+4\eta^{2}\left(h^{T}\hat{h}-2h^{T}\hat{y}+y^{T}y\right)hh^{T}\\
 & = & Q-2\eta\left(h\hat{z}^{T}+\hat{z}h^{T}\right)+(4\eta^{2}L)hh^{T}
\end{eqnarray*}

which is the update listed as step 9 in the online algorithm (Algorithm
\ref{alg:online} \vpageref{alg:online}).

\newpage{}

\section{Details regarding controlling numerical stability}

The update of $U$ (step 6 of the online algorithm, step 7 in the
minibatch version) may over time lead to $U$ becoming ill-conditioned.
Simultaneously, as we update $U$ and $U^{-T}$ (using Sherman-Morrison
or Woodbury) our updated $U^{-T}$may numerically start to diverge
from the true $U^{-T}$due to numerical precision. It is thus important
to prevent both of these form happening, i.e. make sure $U$ stays
well conditioned, to ensure the numerical stability of the algorithm.
We present here progressively refined strategies for achieving this.

\subsection*{Restoring the system in a pristine stable state}

One simple way to ensure numerical stability is to once in a while
restore the system in its pristine state where $V=W$ and $U=\mathbf{I}_{d}=U^{-T}$.
This is easily achieved as follows: 
\begin{eqnarray*}
V & \leftarrow & VU\\
U & \leftarrow & \mathbf{I}_{d}\\
U^{-T} & \leftarrow & \mathbf{I}_{d}.
\end{eqnarray*}
This operation doesn't affects the product $VU$, so the implicit
matrix $W$ remains unchanged, nor does it affect $Q=W^{T}W$. And
it does restore $U$ to a perfectly well conditioned identity matrix.
But computing $VU$ is an extremely costly $O(Dd^{2})$ operation,
so if possible we want to avoid it (except maybe once at the very
end of training, if we want to compute the actual $W$). In the next
paragraphs we develop a more efficient strategy.

\subsection*{Stabilizing only problematic singular values}

$U$ becoming ill-conditioned is due to its singular values over time
becoming too large and/or too small. Let use define $\sigma_{1},~\ldots,~\sigma_{d}$
as the singular values of $U$ ordered in decreasing order. The conditioning
number of $U$ is defined as $\frac{\sigma_{1}}{\sigma_{d}}$ and
it can become overly large when $\sigma_{1}$ becomes too large and/or
when $\sigma_{d}$ becomes too small. Restoring the system in its
pristine state, as shown in the previous paragraph, in effect brings
back \emph{all} singular values of $U$ back to 1 (since it brings
back $U$ to being the identity). It is instead possible, and computationally
far less costly, to correct when needed only for the singular values
of $U$ that fall outside a safe range. Most often we will only need
to occasionally correct for one singular value (usually the smallest,
and only when it becomes too small). Once we have determined the offending
singular value and its corresponding singular vectors, correcting
for that singular value, i.e. effectively bringing it back to 1, will
be a $O(Dd)$ operation. The point is to apply corrective steps only
on the problematic singular values and only when needed, rather than
blindly, needlessly and inefficiently correcting for all of them through
the basic $O(Dd^{2})$ full restoration explained in the previous
paragraph.\pagebreak{}

Here is the detailed algorithm that achieves this:

\begin{algorithm}[H]
\protect\caption{Numerical stabilization procedure for problematic singular values}

\begin{itemize}
\item The chosen safe range for singular values is $[\sigma_{\mathrm{low}}, \sigma_{\mathrm{high}}]$ (ex: $[0.001, 100]$ )
\item The procedures given below act on output layer parameters $U$, $U^{-T}$ and $V$. 
\item For concision, we do not enlist these parameters explicitly in their parameter list.
\item Procedure \textsc{singular-stabilize} gets called after every $n_{\mathrm{check}}$ gradient updates (ex: $n_{\mathrm{check}}=100$).
\end{itemize}

\begin{algorithmic} 

\Procedure{singular-stabilize}{$\,$}
  \State $\bar{\mathbf{U}}$, $\sigma$, $\bar{\mathbf{V}}$ = \Call{SVD}{$U$} \Comment{Computes singular value decomposition of $U$ as $U = \bar{\mathbf{U}}\, \mathrm{diag}(\sigma)\, \bar{\mathbf{V}}^T $}
  \ForAll{ $k \in \{1, \ldots, d \}$ }
    \If{$\sigma_k$ < $\sigma_{\mathrm{low}}$ OR $\sigma_k$ > $\sigma_{\mathrm{high}}$}
      \State \Call{fix-singular-value}{$\sigma_k$, $\bar{\mathbf{U}}_k$, $1$}
    \EndIf
  \EndFor
\EndProcedure

\vspace*{0.5cm}

\noindent \emph{The following procedure will change singular value $\sigma$ of $U$ associated to singular vector $u$ to become target singular value $\sigma^*$ (typically 1). It doesn't change $U$'s singular vectors, only that one singular value. It also changes $V$ symetrically (with a rank-one update) in such a way that $W=VU$ remains unchanged.}
\vspace*{0.5cm}

\Procedure{fix-singular-value}{$\sigma$, $u$, $\sigma^*$} 
\State $ \alpha = \frac{\sigma^*-\sigma}{\sigma}$
\State $ \beta = - \frac{\alpha}{1+\alpha}$
\State $U \leftarrow U + \alpha u (U^T u)^T$
\State $V \leftarrow V + \beta (V u) u^T $
\State $U^{-T} \leftarrow U^{-T} + \beta u (U^{-1} u)^T$ \Comment{Where $U^{-1}$ is obtained as the transpose of $U^{-T}$. But we may instead of this prefer to recompute $U^{-T}$ from scratch by inverting $U$ to ensure it doesn't stray too much due to numerical imprecisions.}
\EndProcedure

\end{algorithmic}
\end{algorithm}

The proof that the \noun{fix-singular-value} procedure achieves what
it is supposed to is relatively straightforward, and left to the reader.

\subsection*{Avoiding the cost of a full singular-value decomposition}

Computing the SVD of $d\times d$ matrix $U$ as required above, costs
roughly $25d^{3}$ elementary operations (use the so-called \noun{r-svd}
algorithm). But since the offending singular values will typically
be only the smallest or the largest, it is wasteful to compute all
$d$ singular values every time. A possibly cheaper alternative is
to use the power iteration method with $U$ to find its largest singular
value and associated singular vector, and similarly with $U^{-1}$to
obtain the smallest singular value of $U$ (which corresponds to the
inverse of the largest singular value of $U^{-1}$). Each iteration
of the power iteration method requires only $O(d^{2})$ operations,
and a few iterations may suffice. In our experiments we fixed it to
100 power iterations. Also it is probably not critical if the power
iteration method is not run fully to convergence, as correcting along
an approximate offending singular vector direction can be sufficient
for the purpose of ensuring numerical stability. 

With this refinement, we loop over finding the smallest singular value
with the power iteration method, correcting for it to be 1 by calling
\noun{fix-singular-value} if it is too small, and we repeat this until
we find the now smallest singular value to be inside the acceptable
range. Similarly for the largest singular values. 

Note that while in principle we may not need to ever invert $U$ from
scratch (as we provided update formulas of $U^{-T}$ with every change
we make to $U$), it nevertheless proved to be necessary to do so
regularly to ensure $U^{-T}$ doesn't stray too much from the correct
value due to numerical imprecisions. Inverting $U$ using Gaussian-elimination
costs roughly $d^{3}$ operations, so it is very reasonable and won't
affect the computational complexity if we do it no more often than
every $d$ training examples (which will typically correspond to less
than 10 minibatches of size 128). In practice, we recompute $U^{-T}$
from scratch every time before we run this check for singular value
stabilization. 
\end{document}